%% file: main.tex
\documentclass[10pt,twocolumn,letterpaper]{article}

\usepackage{cvpr}              
\usepackage{times}
\usepackage{bm}
\usepackage{microtype}
\usepackage{epsfig}
\usepackage{caption}
\usepackage{float}
\usepackage{placeins}
\usepackage{color, colortbl}
\usepackage{stfloats}
\usepackage{enumitem}
\usepackage{tabularx}
\usepackage{xstring}
\usepackage{multirow}
\usepackage{xspace}
\usepackage{url}
\usepackage[hang,flushmargin]{footmisc}
\usepackage{algorithmic}
\usepackage{algorithm}
\usepackage{footnote}
\input{preamble}

\usepackage[accsupp]{axessibility}
\definecolor{cvprblue}{rgb}{0.21,0.49,0.74}
\usepackage[pagebackref,breaklinks,colorlinks,citecolor=cvprblue]{hyperref}
\usepackage{amsthm,amsmath,amssymb}
\usepackage{hyperref}
\usepackage{mathrsfs}

\def\paperID{1614} 
\def\confName{CVPR}
\def\confYear{2025}

\title{MagicFace: Training-free Universal-Style Human Image Customized Synthesis}

\author{
	\fontsize{12}{14}\selectfont
	\textbf{Yibin Wang}\textsuperscript{1},
	\textbf{Weizhong Zhang}\textsuperscript{2}\footnotemark[2],
	\textbf{Cheng Jin}\textsuperscript{1}\footnotemark[2] \\
	\textsuperscript{1}School of Computer Science, Fudan University 
        \textsuperscript{2}School of Data Science, Fudan University \\
	{\tt\small yibinwang1121@163.com, weizhongzhang@fudan.edu.cn, jc@fudan.edu.cn}\\
    \href{https://codegoat24.github.io/MagicFace/}{codegoat24.github.io/MagicFace}
}

\begin{document}

\twocolumn[{%
	\renewcommand\twocolumn[1][]{#1}%
	\maketitle
	\begin{center}
		\centering
		\captionsetup{type=figure}
		\includegraphics[width=0.95\linewidth]{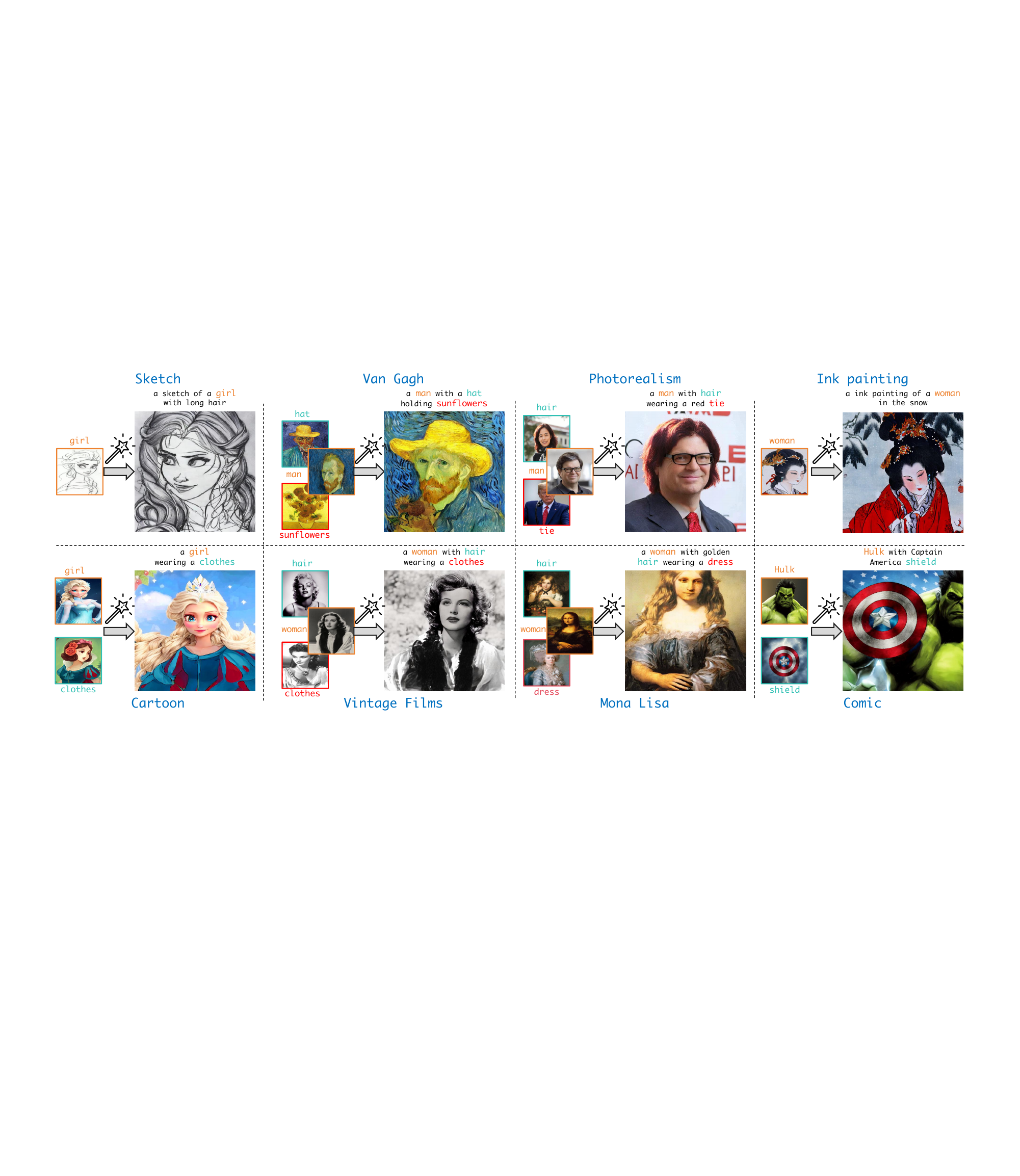}
		\captionof{figure}{\textbf{Results of single/multi-concept customization for humans across diverse styles.} Each instance comprises two distinct inputs: a textual description, and a set of reference concept images. MagicFace is capable of generating high-fidelity images depicting the specific individual of any style with multiple given concepts in a training-free manner. Best viewed on screen.}
		\label{fig:display}
	\end{center}
}]

\input{00_abstract}

\input{01_intro}
\input{figs/model1}

\input{02_related}

\input{03_method}

\input{04_experiment}

\input{10_conclusion}
{
    \small
    \bibliographystyle{ieeenat_fullname}
    \bibliography{11_references}
}

 \clearpage \input{12_appendix}

\end{document}

%% file: preamble.tex
\usepackage[dvipsnames]{xcolor}


%% file: 00_abstract.tex
\begin{abstract}
Current human image customization methods leverage Stable Diffusion (SD) for its rich semantic prior. 
However, since SD is not specifically designed for human-oriented generation, these methods often require extensive fine-tuning on large-scale datasets, which renders them susceptible to overfitting and hinders their ability to personalize individuals with previously unseen styles.
Moreover, these methods extensively focus on single-concept human image synthesis and lack the flexibility to customize individuals using multiple given concepts, thereby impeding their broader practical application.
This paper proposes MagicFace, a novel training-free method for multi-concept universal-style human image personalized synthesis. 
Our core idea is to simulate how humans create images given specific concepts, i.e., first establish a semantic layout considering factors such as concepts' shape and posture, then optimize details by comparing with concepts at the pixel level. To implement this process, we introduce a coarse-to-fine generation pipeline, involving two sequential stages: semantic layout construction and concept feature injection. This is achieved by our Reference-aware Self-Attention (RSA) and Region-grouped Blend Attention (RBA) mechanisms.  
In the first stage, RSA enables the latent image to query features from all reference concepts simultaneously, extracting the overall semantic understanding to facilitate the initial semantic layout establishment. 
In the second stage, we employ an attention-based semantic segmentation method to pinpoint the latent generated regions of all concepts at each step. Following this, RBA divides the pixels of the latent image into semantic groups, with each group querying fine-grained features from the corresponding reference concept.
Notably, our method empowers users to freely control the influence of each concept on customization through a weighted mask strategy.
Extensive experiments demonstrate the superiority of MagicFace in both single- and multi-concept human image customization. 

\end{abstract}

%% file: 01_intro.tex
\section{Introduction}
\label{sec:intro}
Text-to-image generation has undergone remarkable advancements with the emergence of large-scale text-to-image diffusion models \cite{sd,wang2024high,primecomposer}. 
One of the most popular and challenging topics in these developments is the person-centric subject-to-image generation, which aims to personalize individuals in novel scenes, styles, and actions given a few reference images. It has garnered significant attention due to its various applications, such as AI portrait photos \cite{wang2024instantid}, image animation \cite{xu2024magicanimate}, and virtual try-ons \cite{xu2024ootdiffusion}, each of which offers unique benefits and practical uses. 

\input{figs/subject-to-image_compare1}
Current methods are extensively based on Stable Diffusion (SD) \cite{sd} to utilize its rich semantic prior. However, since SD is not specifically designed for human generation, extensive fine-tuning is typically required, which takes on two main approaches: either tuning-based or zero-shot customization.
The former methods such as Dreambooth \cite{dreambooth}, aim to fine-tune pre-trained text-to-image models with dozens of reference images during testing to better reflect the new concepts. 
The latter methods train visual encoders \cite{chen2023photoverse,photomaker,wang2024instantid} or hypernetworks \cite{arar2023domain,ruiz2024hyperdreambooth,han2024face} on existing human-centric datasets. These components represent the reference images as embeddings or LoRA \cite{hu2021lora} weights, integrating identity features into the diffusion generation process via attention layers. After training, the model is capable of personalizing human images using a single image per individual through a single forward pass.

However, these training-based methods are prone to overfitting \cite{tewel2023key,photomaker} and quality compromises \cite{face-diffuser}. Limited by their training datasets, these methods are mainly effective for personalizing individuals of photorealistic style but struggle to adapt to unseen styles given by users, such as cartoon characters or painted figures. Additionally, these methods are tailored for single-concept human image synthesis and encounter challenges in complex scenarios involving multiple concepts, such as adding given accessories like a tie to a specific individual.
Recent studies on multi-concept customized synthesis \cite{custom-diffusion,purushwalkam2024bootpig,freecustom} seem to offer effective solutions. However, 
these approaches are primarily effective on general objects with coarse-grained textures and perform poorly in human image customization (see \cref{fig:multi-concept_compare}) because facial identity involves more nuanced semantics and demands finer guidance during the customization.

This paper proposes MagicFace, a novel training-free method for multi-concept universal-style human image personalized synthesis.
Instead of reflecting new concepts by fine-tuning model parameters, we simulate the human drawing process for specific concepts: first depict a semantic layout based on concepts' coarse factors like shape and posture, then refine the details by referencing their fine-grained features.
Following this, MagicFace introduces a coarse-to-fine forward generation pipeline involving two sequential stages, semantic layout construction, and concept feature injection, that extract visual features from reference concepts for high-fidelity customized synthesis in an evolving scheme. The core idea lies in our Reference-aware Self-Attention (RSA) and Region-grouped Blend Attention (RBA) mechanisms. Specifically, in the first stage, RSA allows the generated image to simultaneously query features from all concepts, adaptively integrating coarse-grained semantic information at each step. This process extracts a comprehensive semantic understanding to facilitate the initial semantic layout construction progressively. 
Once the overall scene is preliminarily established, we advance to the second stage for concept feature refinement: at each step, we utilize an attention-based semantic segmentation method to identify the latent generated regions of all concepts. To be precise, the cross-attention map is renormalized to assign each patch to the specific semantic concept, while the self-attention map is employed to refine and complete the semantic regions. Based on this, RBA divides the pixels of the latent image into semantic groups and enables each group to query fine-grained features from the corresponding reference concept. 
This process effectively guarantees the precise alignment of attributes and meticulous feature integration for each concept. Besides, 
to encourage the model to focus more on given concepts and eliminate irrelevant information in reference images, we introduce a weighted mask strategy \cite{ohanyan2024zero,freecustom} that rectifies the model's attention. This also allows users to freely adjust each concept's influence on customization. 

To the best of our knowledge, this is the first method to achieve universal-style human image personalized synthesis that enables multi-concept customization without requiring any training. 
Extensive experiments have demonstrated that MagicFace surpasses current state-of-the-art approaches in both human-centric subject-to-image synthesis and multi-concept human image customization. 
It also can be applied to texture transfer, further showcasing its versatility and effectiveness. Our contributions are summarized as follows:
\begin{itemize}
    \item We propose a novel training-free method for high-fidelity human image personalized synthesis, leveraging our RSA and RBA.
    \item Our method enables multi-concept customization for humans of any style. Additionally, it can be utilized for texture transfer, demonstrating its versatility and broad applicability.
    \item Extensive experiments demonstrate the superiority of our proposed method in both human-centric subject-to-image synthesis and multi-concept human image customization.

\end{itemize}

%% file: figs/subject-to-image_compare1.tex
\begin{figure}[t]
    \centering
    \includegraphics[width=1\linewidth]{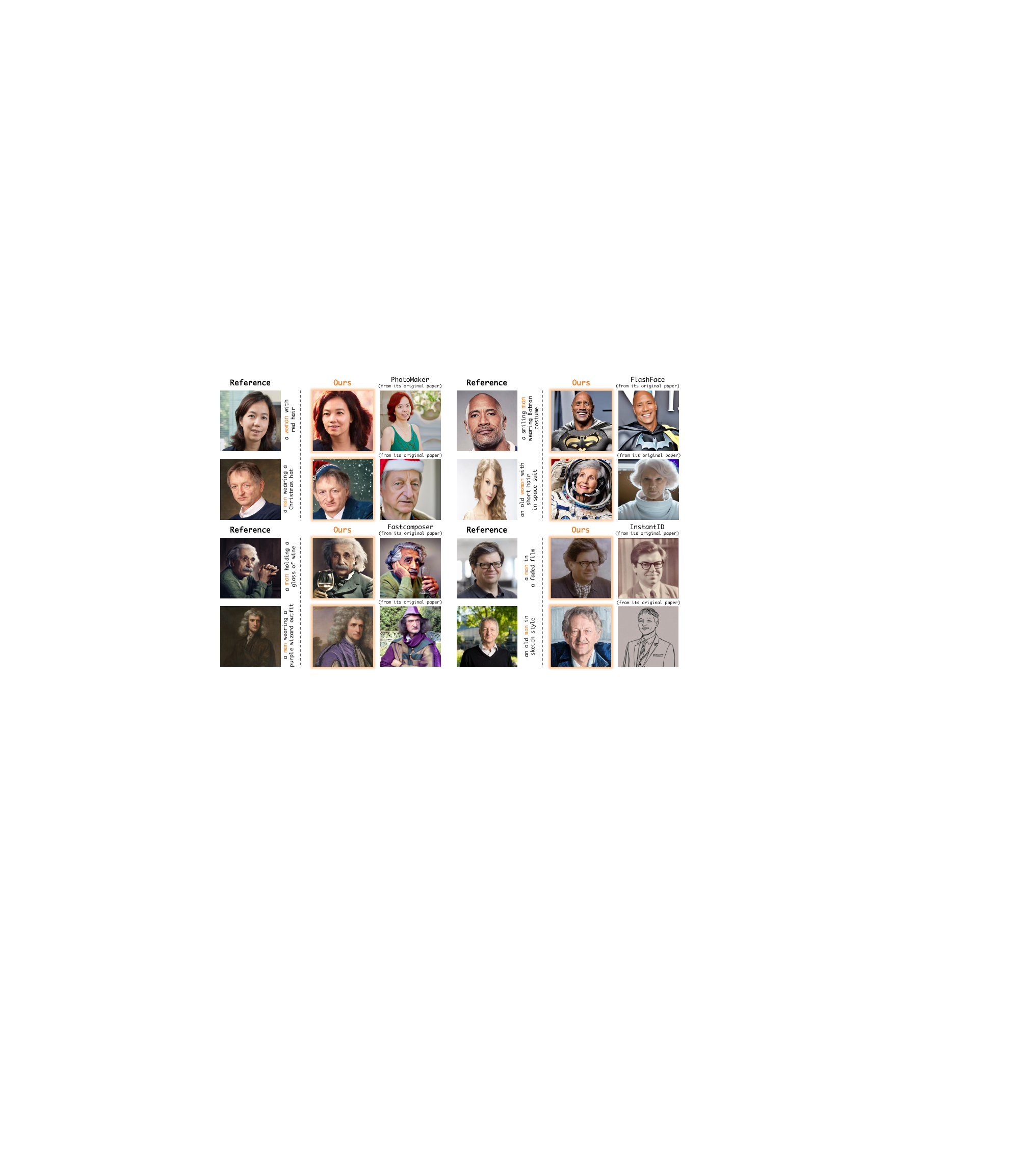}
    \caption{Qualitative comparison on human-centric subject-to-image generation.}
    \label{fig:subject-to-image_compare}
    \vspace{-0.3cm}
\end{figure}

%% file: figs/model1.tex
\begin{figure*}[t]
    \centering
    \includegraphics[width=1\linewidth]{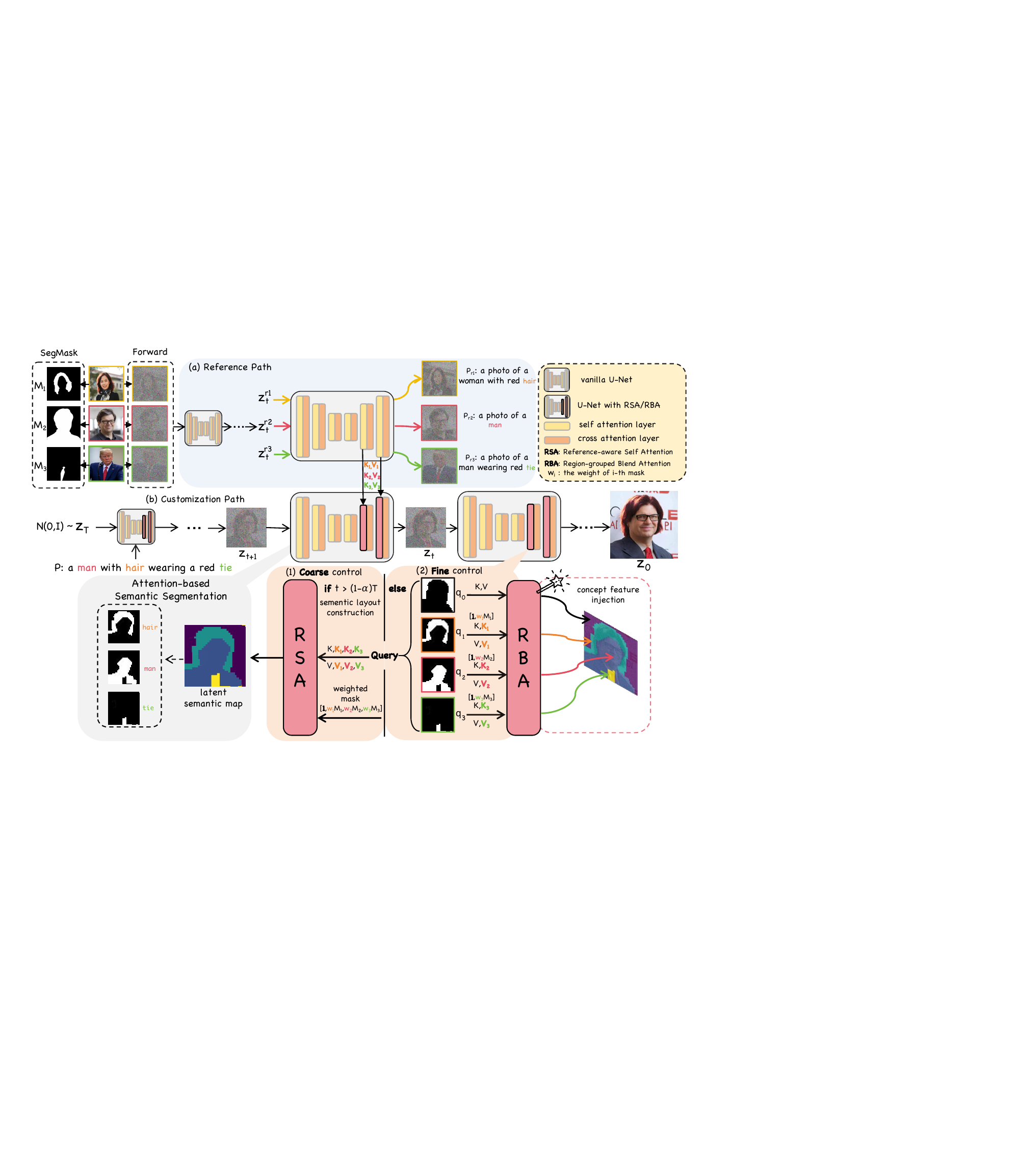}
    \caption{\textbf{Overview of our MagicFace}. Given reference images, their segmentation masks, and text prompts, we generate personalized image $\textbf{\textit{z}}_{0}$ aligned to the target prompt $P$. The sampling pipeline consists of two paths: (a) the reference path and (b) the customization path. 
    In (a), we first employ a diffusion forward process on the reference images. Then, the noised reference latents are input into vanilla U-Net. In (b), we first sample a Gaussian noise $\textbf{\textit{z}}_T$ and introduce a coarse-to-fine generation process involving two sequential stages: semantic layout construction and concept feature injection. At each step $t$, we pass latent $\textbf{\textit{z}}_t$ to our modified U-Net: (1) in the first stage, we employ RSA to integrate the features from the reference path to facilitate the initial semantic scene construction; (2) in the second stage, we first obtain the latent semantic map of $\textbf{\textit{z}}_{t}$ via attention-based segmentation method. Based on this, RBA divides the latent image and ensures fine-grained feature injection for each generated concept. A weighted mask strategy is adopted to ensure the model focuses more on given concepts.}
    \label{fig:model}
\end{figure*}

%% file: 02_related.tex
\section{Related Work}
\label{sec:related}
Current human personalized synthesis methods can be mainly divided into two categories: tuning-based customization and zero-shot customization, which will be briefly discussed in this section. Besides, we will also introduce the related studies on multi-concept image customization.

\textbf{Tuning-based Customization.}
Tuning-based methods rely on additional subject-specific optimization during test time. The pioneering work, Dreambooth \cite{dreambooth}, fine-tunes a text-to-image diffusion model using numerous reference images to bind a unique identifier to the given subject. A concurrent work, Textual Inversion \cite{textual_inversion} transforms subject images into a simple learnable text embedding to encode the subject's identity. Following this, subsequent works such as NeTI \cite{alaluf2023neural} and XTI \cite{XTI} introduce implicit time-aware representation and layer-wise learnable embedding respectively to achieve better performance. Additionally, Tuning-Encoder \cite{tuning-encoder} generates an initial set of latent codes using a pre-trained encoder and refines these codes with minimal fine-tuning iterations to better preserve subject identities.
Despite their effectiveness, these methods always encounter inefficiency due to their demand of considerable time and computing resources for fine-tuning in the test time. 

\textbf{Zero-shot Customization.}
Zero-shot methods attempt to perform customization using a single image with a single forward pass, significantly accelerating the personalization process. For example, 
ELITE \cite{wei2023elite} and InstantBooth \cite{shi2023instantbooth} achieve this by utilizing a global mapping network to encode reference images into word embeddings and a local mapping network to inject reference image patch features into cross-attention layers. Fastcomposer \cite{fastcomposer} and PhotoMaker \cite{photomaker} extract identity-centric embeddings by fine-tuning the image encoder and merging the class and image embeddings. 
Face-Diffuser \cite{face-diffuser} reveals the training imbalance and quality compromise issues in these methods and addresses them by proposing a novel collaborative generation pipeline. InstantID \cite{wang2024instantid} proposes an IdentityNet to steer the generation by integrating facial and landmark images with textual prompts. FlashFace \cite{flashface} encodes the reference image into a series of feature maps to maintain more details.
Despite their impressive results, these methods are struggle to personalize individuals of unseen styles limited by their training datasets.
Besides, they only focus on single-concept human personalization and fail to deal with complex scenarios involving multiple concepts.

\textbf{Multi-concept Image Customization.}
Different from these works, recent studies focus on multi-subject customization. Custom Diffusion \cite{custom-diffusion} achieves this by closed-form constrained optimization for multiple concepts. Perfusion \cite{tewel2023key} proposes a dynamic rank-1 update strategy to maintain high visual fidelity. FreeCustom \cite{freecustom} proposes a multi-reference self-attention mechanism, allowing the latent image to interact with the input concepts. ClassDiffusion \cite{huang2024classdiffusion} leverages a semantic preservation loss to explicitly regulate the concept space when learning a new concept. Despite their effectiveness on general objects with coarse-grained textures, these approaches fall short in human customization due to the nuanced semantics and higher level of detail and fidelity required for facial identity.

%% file: 03_method.tex
\section{MagicFace}
\label{sec:method}

\subsection{Overview}
We present an overview of our MagicFace in \cref{fig:model}. Our goal is to achieve high-fidelity human image personalized synthesis, enabling multi-concept customization for humans across various styles. To accomplish this, we mimic the human painting process and propose a coarse-to-fine generation pipeline. 

The overall pipeline involves two paths: the reference path and the customization path. Since our method is entirely training-free, it operates as a standard Stable Diffusion process, ensuring inherent alignment between the two paths. This alignment allows the reference path to seamlessly supply the customized path with essential prior information from the reference image throughout the generation process. The reference path is relatively straightforward, primarily managed by Stable Diffusion. Our primary technical innovations are implemented in the customization path to enable high-fidelity human image personalized generation. 
Specifically, the sampling process in this path consists of two sequential stages involving $T$ denoising steps.
\begin{itemize}
 \item In the first stage, we employ RSA to extract the overall semantic understanding from all given concepts, facilitating the initial semantic scene construction (\cref{sec:first_stage}). 

\item In the second stage, a latent semantic map is first derived to pinpoint the generated regions of all concepts at each step. We then utilize RBA to accurately inject features of reference concepts into their corresponding regions (\cref{sec:second_stage}), ensuring the generated concepts closely resemble the reference images. 
\end{itemize}
To emphasize the model's attention to the reference concepts, a weighted mask strategy is implemented during customized synthesis (\cref{sec: weighted_mask}).


\subsection{Coarse-to-Fine Generation Process}
MagicFace synthesizes images by progressing from coarse-grained semantic scene construction with $\alpha T$  steps to fine-grained feature injection with $T$(1-$\alpha$) steps in an evolving scheme. 
In this section, we will first elaborate on the two paths in the pipeline and then delve into the details of each sampling stage.


\textbf{Reference Path.}
For each concept image in the reference path, we first apply a diffusion forward process \cite{ddpm} to compute the noised reference latent $\textbf{\textit{z}}_{T}^{'}$. At each step $t$, we input $\textbf{\textit{z}}_{t}^{'}$ and its text prompt into the U-Net $\epsilon_\theta$. 
In the self-attention layer $l$ and time step $t$,  
we extract the key $\textbf{\textit{K}}_{i,l,t}$ and value $\textbf{\textit{V}}_{i,l,t}$ of $i$-th concept to guide the synthesis in the customization path.

\textbf{Customization Path.} 
This path starts by sampling the latent $\textbf{\textit{z}}_T$ from the Gaussian distribution. Then,
we modify the vanilla self-attention module in SD U-Net, extending it into our RSA/RBA. At each step $t$, we pass the latent $\textbf{\textit{z}}_t$ and target prompt $P$ into modified U-Net $\epsilon_\theta^*$, integrating concept features from $\textbf{\textit{K}}_{i,l,t}$ and $\textbf{\textit{V}}_{i,l,t}$ derived in reference path. The final denoised result is the ultimate synthesized image.

In the following, we will focus on two sampling stages in the customization path.

\subsubsection{Semantic Scene Construction and Segmentation} \label{sec:first_stage}
\textbf{Reference-aware Self-Attention.}
RSA enables the latent image to query features from all reference concepts simultaneously to integrate coarse-grained semantic information for initial semantic scene construction. Specifically, in each self-attention layer $l$ and time step $t$, the query, key, and value feature from $\textbf{\textit{z}}_t$ are $\textbf{\textit{Q}}_{l,t}$, $\textbf{\textit{K}}_{l,t}$ and $\textbf{\textit{V}}_{l,t}$. For brevity, we omit $l, t$ in the following. We then concatenate $\textbf{\textit{K}}$ and $\textbf{\textit{V}}$ with injected key and value features of $N$ concepts from reference path, resulting in $\hat{\textbf{\textit{K}}}$ = $[\textbf{\textit{K}}, \textbf{\textit{K}}_1, \textbf{\textit{K}}_2,..., \textbf{\textit{K}}_N]$ and $\hat{\textbf{\textit{V}}}$ = $[\textbf{\textit{V}}, \textbf{\textit{V}}_1, \textbf{\textit{V}}_2,..., \textbf{\textit{V}}_N]$. However, we aim to query features exclusively from the concept region, as irrelevant background information in the reference image may distract the model and reduce its effectiveness. To address this issue, we rectify the model's attention by introducing the segmentation mask $\textbf{\textit{M}}_i$ of each reference concept and concatenate them with an all-ones matrix to obtain $\textbf{\textit{M}}$ = $[\textbf{1}, w_1\textbf{\textit{M}}_1, w_2\textbf{\textit{M}}_2,..., w_N\textbf{\textit{M}}_N]$ where $w_i$ represents the scaling weight for each mask (elaborated in \cref{sec: weighted_mask}). Ultimately, the RSA can be formulated as follows:
\begin{equation}
    \text{RSA}(\textbf{\textit{Q}},\hat{\textbf{\textit{K}}},\hat{\textbf{\textit{V}}}, \textbf{\textit{M}})=\text{Softmax}(\frac{\textbf{\textit{M}} \odot \textbf{\textit{Q}}\hat{\textbf{\textit{K}}}^\mathrm{T}}{\sqrt{d}})\hat{\textbf{\textit{V}}}.
\end{equation}
Here, $\odot$ denotes the Hadamard product. This process ensures the generated image effectively interacts with the overall semantic information from the reference images while filtering out unrelated noise.

\noindent\textbf{Latent Semantic Map Generation.}
To ensure precise feature injection for each concept at the pixel level, we need to identify their latent generated region. However, this task is challenging because the latent image at each step is not accessible during the generation process. Fortunately, the attention layers in the U-Net contain rich semantic information, which can be utilized to identify semantic units efficiently \cite{chefer2023attend}. 
Therefore, we compute the latent semantic map based on attention maps, involving two sequential steps: cross-attention-based semantic segmentation and self-attention-based segmentation completion.

\textbf{Attention Maps Generation.}
In the attention layer $l$ and time step $t$, a self-attention map $\textbf{\textit{S}}_{t}^{l}$ and a cross-attention map $\textbf{\textit{C}}_{t}^{l}$ are calculated over linear projections of the intermediate image spatial feature $\textbf{\textit{z}}_t^l$ or text embedding $\textbf{\textit{e}}$, 
\begin{align}
\textbf{\textit{S}}_{t}^{l}=\text{Softmax}\left(\frac{\textbf{\textit{Q}}_s(\textbf{\textit{z}}_t^l)\textbf{\textit{K}}_s(\textbf{\textit{z}}_t^l)^\mathrm{T}}{\sqrt{d}}\right),\\
    \textbf{\textit{C}}_{t}^{l}=\text{Softmax}\left(\frac{\textbf{\textit{Q}}_c(\textbf{\textit{z}}_t^l)\textbf{\textit{K}}_c(\textbf{\textit{e}})^\mathrm{T}}{\sqrt{d}}\right),
\end{align}
where $\textbf{\textit{Q}}_\ast$(·) and $\textbf{\textit{K}}_\ast$(·) are linear projections with the dimension $d$. Though self-attention modules are modified in our framework, we still compute the original self-attention maps for latent semantic map generation at each step.

\textbf{Semantic Segmentation.} Cross-attention maps contain the similarity values between image patch \textit{s} and text token $P_i$. Therefore, in each row of $\textbf{\textit{C}}_{t}^{l}$, a higher probability $\textbf{\textit{C}}_{t}^{l}[s, i]$ indicates a closer relationship between $s$ and $P_i$. Based on this, we segment the $\textbf{\textit{z}}_{t}$ as the set of regions masked by $[\textbf{\textit{m}}_1,\textbf{\textit{m}}_2,...,\textbf{\textit{m}}_K]$, with $K$ denotes the number of tokens and $\textbf{\textit{m}}_i \in \{0,1\}$ indicate the latent semantic region of token $P_i$.

Specifically, we first upsample all cross-attention maps $\textbf{\textit{C}}_t^l$ to the same size, then average and renormalize them along the spatial dimension to obtain the final cross-attention map, formulated as $\textbf{\textit{C}}_{t} = \Phi(\textbf{\textit{C}}_t^l)$. This map estimates the likelihood of assigning patch $s$ for the token $P_i$. Ultimately, the argmax operation is applied to the token dimension to determine the activation of each patch:
\begin{align}
i_s=\arg\max_i{\textbf{\textit{C}}}_t[s,i].
\label{equ:seg}
\end{align}
Along this line, we compute $\textbf{\textit{m}}_i$ by setting the element in the patch set $\{s:i_s = i\}$ as 1, and others to 0. Note that we only need the semantic masks of $N$ reference concepts for feature injection. Hence, we retain the masks of concept-specific tokens and merge the remaining ones into a new mask $\textbf{\textit{m}}_0$ indicating the background region, yielding $\mathcal{M}=[\textbf{\textit{m}}_0, \textbf{\textit{m}}_1, \textbf{\textit{m}}_2,...,\textbf{\textit{m}}_N]$. 

\textit{Visualization.} The first row in \cref{fig:seg_example} demonstrates the result of the aforementioned semantic segmentation results. The semantic maps effectively identify the rough locations of the generated concepts. However, they often exhibit unclear boundaries and may contain internal gaps, leading to problematic results. To address this issue, we refine and complete the semantic map using self-attention maps \cite{wang2023diffusion}, as detailed in the following section.

\textbf{Segmentation Completion.}
Self-attention maps $\textbf{\textit{S}}_t^l$ estimate the correlation between image patches, thereby can be used to refine incomplete activation regions in cross-attention maps via transmitting semantic information across patches. This approach is analogous to the feature propagation in spectral graph convolution \cite{kipf2016semi}, as self-attention maps can be viewed as transition matrices where each element is nonnegative and the sum of each row sums to 1. Consequently, we refine the cross-attention maps $\textbf{\textit{C}}_t^l$ by multiplying them with their corresponding self-attention maps, yielding $\hat{\textbf{\textit{C}}}_t^l=\textbf{\textit{S}}_t^l\textbf{\textit{C}}_t^l$. Based on this, the refined final cross-attention map denoted as $\hat{\textbf{\textit{C}}}_t = \Phi(\hat{\textbf{\textit{C}}}_t^l)$, is then used in \cref{equ:seg} to derive refined latent semantic masks $\mathcal{M}$.
\input{figs/segmentation_example1}


\textit{Visualization.} The last row in \cref{fig:seg_example} demonstrates the effectiveness of the refinement where the resulting segmentation maps indicate clearer object boundaries and fewer internal holes.

\subsubsection{Concept Feature Injection} \label{sec:second_stage}
\textbf{Region-grouped Blend Attention.}
With the latent semantic masks $\mathcal{M}$, RBA divides the latent image into concept groups, enabling each generated concept to query fine-grained features from the corresponding reference image. Specifically, at each self-attention layer $l$ and time step $t$, the query, key, and value feature from $\textbf{\textit{z}}_t$ are $\textbf{\textit{Q}}$, $\textbf{\textit{K}}$, and $\textbf{\textit{V}}$. We omit $l,t$ for brevity. Next, we segment the $\textbf{\textit{Q}}$ into groups $[\textbf{\textit{q}}_0, \textbf{\textit{q}}_1,..., \textbf{\textit{q}}_N]$ based on $\mathcal{M}$. 
For each group $\textbf{\textit{q}}_i$ where $i>0$, we concatenate $\textbf{\textit{K}}$ and $\textbf{\textit{V}}$ with their corresponding injected key and value features from reference path, resulting in $\Tilde{\textbf{\textit{K}}}_i=[\textbf{\textit{K}}, \textbf{\textit{K}}_i]$ and $\Tilde{\textbf{\textit{V}}}_i=[\textbf{\textit{V}}, \textbf{\textit{V}}_i]$. For the feature group $\textbf{\textit{q}}_0$ corresponding to the background, we have $\Tilde{\textbf{\textit{K}}}_0=\textbf{\textit{K}}$ and  $\Tilde{\textbf{\textit{V}}}_0=\textbf{\textit{V}}$. 
Based on this, we individually compute the region-grouped attention output $\textbf{\textit{x}}_i$ as:
\begin{equation}
    \textbf{\textit{x}}_i=\text{Softmax}(\frac{\Tilde{\textbf{\textit{M}}}_i \odot \textbf{\textit{q}}_i\Tilde{\textbf{\textit{K}}}_i^\mathrm{T}}{\sqrt{d}})\Tilde{\textbf{\textit{V}}}_i,
\end{equation}
where $\Tilde{\textbf{\textit{M}}}_i = [\textbf{1}, w_i\textbf{\textit{M}}_i]$ for $i > 0$, and $\Tilde{\textbf{\textit{M}}}_0 $ is an all-ones matrix. Finally, we blend the outputs $[\textbf{\textit{x}}_0, \textbf{\textit{x}}_1,...,\textbf{\textit{x}}_N]$ into a single output by putting the pixels from each output into their corresponding positions based on $\mathcal{M}$. This process effectively ensures
precise attribute alignment and feature injection for each generated concept.

\subsection{Weighted Mask Strategy} \label{sec: weighted_mask}
Despite segmentation masks of reference concepts helping to mitigate the disturbance of unrelated background information, the model still struggles to accurately capture the unique attributes of the target concept, particularly when it comes to preserving human identity. Therefore, we introduce a scaling factor $w_i$ to each mask $\textbf{\textit{M}}_i$ during the RSA and RBA processes, aiming to enhance the model's focus on the desired concept features. The effectiveness of this strategy is examined in \cref{fig:weighted_mask} and further discussed in \cref{sec:choice_weights}.

\input{figs/weight_choice1}
\input{figs/visualized_attention1}

%% file: figs/segmentation_example1.tex
\begin{figure}[htb]
    \centering
    \includegraphics[width=1\linewidth]{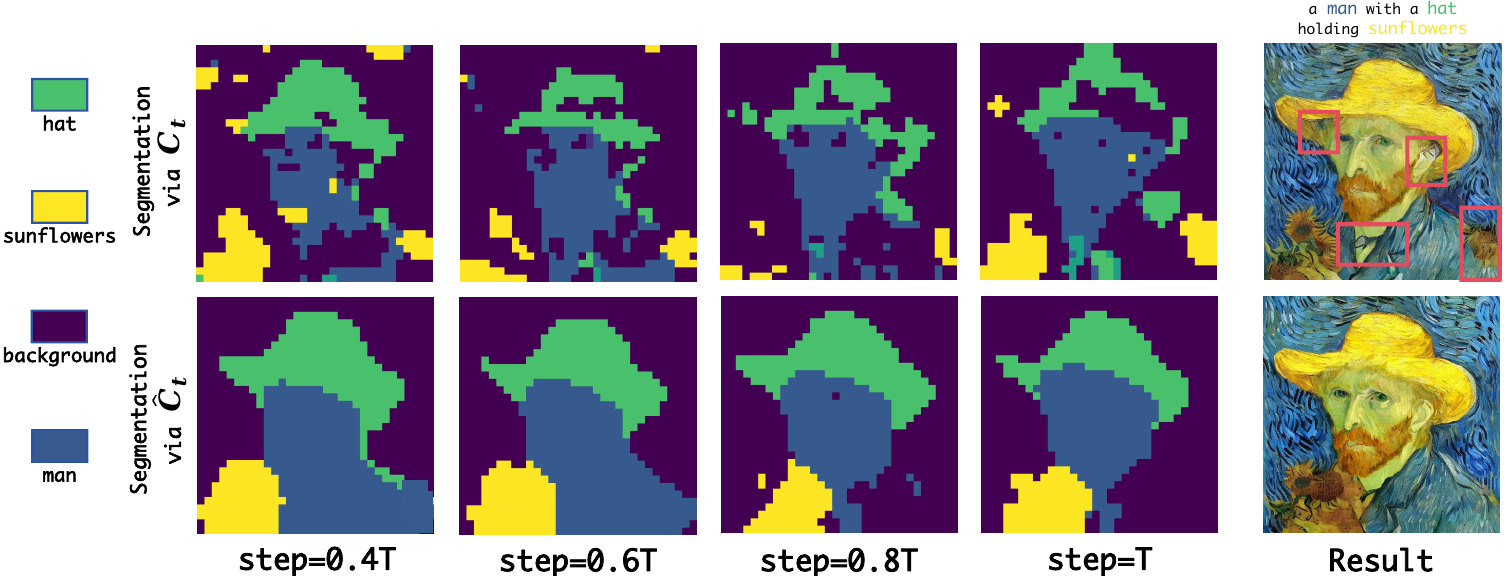}
    \caption{\textbf{Visualization of semantic maps derived by}  $\textbf{\textit{C}}_t$ \textbf{and} $\hat{\textbf{\textit{C}}}_t$. Regions labeled with different colors correspond to different concepts. Problematic regions are highlighted using the outline.}
    \label{fig:seg_example}
\end{figure}

%% file: figs/weight_choice1.tex
\begin{figure}[bht]
    \centering
    \includegraphics[width=1\linewidth]{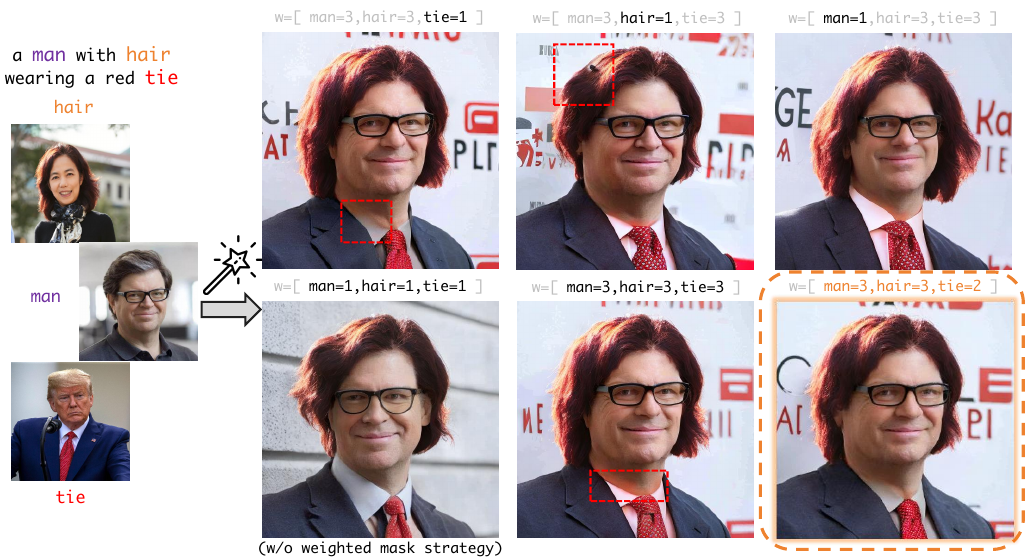}
    \caption{\textbf{Visualized results under different weight settings} $w$. Problematic regions are highlighted using the outline.}
    \label{fig:weighted_mask}
    \vspace{-0.2cm}
\end{figure}

%% file: figs/visualized_attention1.tex
\begin{figure}[htb]
    \centering
    \includegraphics[width=1\linewidth]{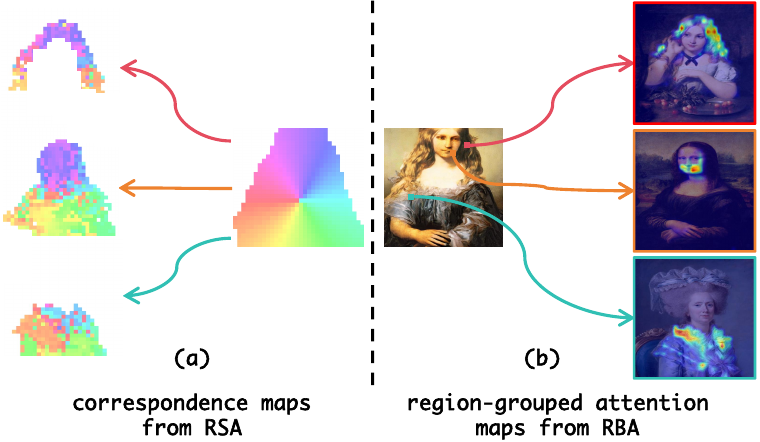}
    \caption{\textbf{Correspondence maps and region-grouped attention maps visualization.} In (a), features with the highest similarity between the generated subject and the reference concepts are marked with the same color. (b) the results of features in colored boxes querying their reference concept keys. }
    \label{fig:visualized_attention}
\end{figure}

%% file: 04_experiment.tex
\section{Experiment}
\label{sec:exp}
\subsection{Implementation Details}
\textbf{Datasets.} 
\textit{Subject-to-Image Generation:} The evaluation dataset consists of 35 human identities (IDs): 24 from previous work \cite{nitzan2022mystyle,face-diffuser} and 11 newly collected. Each ID includes 3 reference images and 1 target image. We also collect 30 prompts covering expressions, attributes, etc. from \cite{photomaker}.
\textit{Multi-concept Customization:}
The evaluation dataset consists of 30 sample images collected from the previous works \cite{freecustom,alaluf2023neural,face-diffuser} and the Internet. Each image includes an ID and 2-3 reference accessories. The segmentation masks of all reference concepts are generated by segment-anything \cite{kirillov2023segment}.

\noindent\textbf{Configurations.} 
Our method is training-free and employs Stable-Diffusion v1.5 as the base model for evaluation using an NVIDIA 3090 GPU. We use 50 steps of the DDIM sampler \cite{ddim}. The scale of classifier-free guidance is set to 7.5. The mask weight $w$ is set to 3 for each concept. We only modify the self-attention module in blocks 5 and 6 of the U-Net. The hyperparameter $\alpha$ is set to 0.4.

\noindent\textbf{Evaluation.} 
We use CLIP-I \cite{gal2022image} and DINO \cite{caron2021emerging} metrics to measure image fidelity, and CLIP-T \cite{radford2021learning} to assess prompt consistency. Face similarity is calculated by detecting facial regions with MTCNN \cite{zhang2016joint} and then computing pairwise ID similarity using FaceNet \cite{schroff2015facenet}.
\input{tables/compare_table}

\subsection{Results}
For a comprehensive evaluation, we compare our method with baselines in human-centric subject-to-image generation and multi-concept customization domains. The quantitative compared results are shown in \cref{table:compare_methods}. 
Notably, our method outperforms FlashFace by 3.2 on the CLIP-T metric. This improvement is expected, as these subject-to-image generation methods, trained on large-scale datasets, tend to overfit and consequently struggle with generating unseen semantic scenes. 
For multi-concept customization, all competing methods exhibit poor performance on the face similarity metric, highlighting their significant challenges in preserving facial identity. 

The qualitative compared results are provided in \cref{fig:subject-to-image_compare} and \cref{fig:multi-concept_compare}. As shown in \cref{fig:subject-to-image_compare}, existing subject-to-image generation methods like FastComposer struggle to achieve personalized generation for the painted character (last row in \cref{fig:subject-to-image_compare}), highlighting their difficulty in customizing humans of unseen styles. Besides, most multi-concept customization methods fail to generate high-fidelity human faces and struggle with complex concepts like neglect (last row in \cref{fig:multi-concept_compare}) as well. Our method produces results that are comparable to, or even more natural and realistic than existing approaches, highlighting the superior human image customized generation capabilities of our method.

\input{figs/multi-concept_compare1}

\subsection{User Study}
We conduct user studies to make a more comprehensive comparison. Specifically, we invite 30 participants through voluntary participation and assign them the task of completing 40 ranking questions. Each question includes reference images, the corresponding text prompt, and the generated images of competitive methods. We anonymize the names of all methods and randomize the order of methods in each question.
The ranking criteria comprehensively considered factors including ID Fidelity, Image Quality, and Prompt Consistency. We collect 23 valid questionnaires for subject-to-image generation and 26 for multi-concept customization.
The results are presented in \cref{tab:user_study_subject} and \cref{tab:user_study_multi}. Notably, our method received favorable feedback from the majority of participants.

\subsection{Ablation Studies} 

\noindent\textbf{Visualization of RSA and RBA.}  
We visualize the correspondence between each feature in the generated image and concept images based on their attention maps in RSA, as well as the attention maps derived by selected image patches querying their corresponding reference concepts in RBA. In \cref{fig:visualized_attention}, (a) the similar regions in generated and reference images indicate higher correlation, and (b) these patches precisely capture features from their reference concepts.

\noindent\textbf{Choice of weights $w$.} \label{sec:choice_weights}
We explore the impacts of different weight settings in \cref{fig:weighted_mask}. We observe that the results improved significantly when adopting the weighted masks.
Although higher weights for each concept lead to more faithful generation, as shown in the first row, excessively high weights can cause coherence issues for easily generated objects such as the tie in this case. Therefore, balancing the weights is crucial to ensure that the generated concepts are both accurate and harmoniously integrated within the scene.

\input{figs/hyper_analysis1}

\noindent\textbf{Choice of the hyperparameter $\alpha$.}
In this work, RSA handles semantic layout construction for $\alpha T$ steps, and RBA takes over for concept feature injection in the remaining steps. We analyze the optimal value of $\alpha$ in \cref{fig:hyper_analysis}. When $\alpha=0$, the generated concepts differ significantly from the reference concepts on shape, such as facial form, due to the absence of RSA. Conversely, when $\alpha=1$, which excludes RBA, the results struggle with identity preservation. These observations highlight the effectiveness of both RSA and RBA. Finally, we find that $\alpha=0.4$ achieves the best balance between semantic layout and identity preservation.

\input{tables/user_study_subject}

\subsection{Applications}

\textbf{Universal-style human customization.}
MagicFace excels in personalizing humans across diverse styles as shown in \cref{fig:display}. Unlike existing methods, which are limited to the photorealistic style due to their training datasets, MagicFace is training-free and effectively customizes humans of different styles by accurately embedding reference image features into the generated concept. To the best of our knowledge, this is the first method to achieve universal-style human image personalized synthesis.

\noindent\textbf{Multi-concept human customization.} 
Compared to current human-centric subject-to-image generation methods which fail to personalize humans with multiple given concepts, our approach enables high-quality multi-concept human customization.
For a more thorough evaluation, we also compare our method against specialized baselines in the multi-concept customization domain. While these methods handle general objects with coarse-grained textures reasonably well, our experimental results indicate that they falter in human-centric customization. In contrast, our approach precisely preserves human identity, setting a new benchmark for multi-concept human image synthesis.

\noindent\textbf{Texture transfer.}
MagicFace also demonstrates high effectiveness in texture transfer, as illustrated in  \cref{fig:appearance_transfer}. By precisely incorporating texture features from input images, it can seamlessly transfer intricate details, such as patterns, colors, and material textures, onto the generated subjects. This capability highlights its versatility in accurately adapting the visual attributes of input images, further showcasing its potential for various customization applications.

\section{Limitation}
First, MagicFace encounters challenges in multi-subject customization due to the inherent limitation of Stable Diffusion. 
When handling tokens with similar or identical semantics, such as "a photo of a \underline{man} and a \underline{man}," Stable Diffusion can cause semantic overlap, also demonstrated by Fastcomposer \cite{fastcomposer}. This overlap misaligns the latent semantic regions associated with concept-specific tokens, leading to incorrect placement of reference concept feature injections and ultimately resulting in identity mixing. Second, MagicFace is capable of synthesizing high-fidelity human images, which, while impressive, raises significant privacy and security concerns. This high level of fidelity can lead to the unauthorized use of personal face portraits, potentially resulting in ethical issues and hindering the broader adoption of such technology. 
\input{tables/user_study_multi}

\input{figs/appearance_transfer1}

\section{Societal Impact}
The societal impact of personalized human image generation technologies, such as MagicFace, is profound. These innovations drive creativity in entertainment, virtual reality, and augmented reality, enabling highly realistic content in video games and films that greatly enhance user experiences. However, as these technologies become increasingly accessible, they raise significant concerns about privacy, consent, and potential misuse. Balancing innovation with ethical considerations is essential to fully realize the benefits of subject-driven text-to-image generation while safeguarding societal interests.

%% file: tables/compare_table.tex
\begin{table*}[ht]  \centering
	\small
        \setlength{\tabcolsep}{2pt}
	\caption{Quantitative comparison against baselines.}
	\label{table1}
	\begin{tabular}{c|cccc|c|cccc}
		\bottomrule
		\multirow{2}{*}{\textbf{Methods}} & \multicolumn{4}{c|}{\textit{\textbf{Subject-to-Image Generation}}} & \multirow{2}{*}{\textbf{Methods}} & \multicolumn{4}{c}{\textit{\textbf{Multi-Concept Customization}}} \\
		  
		& CLIP-T      & CLIP-I     &DINO & Face-sim.     & & CLIP-T      & CLIP-I     &DINO & Face-sim.      \\
		\bottomrule

            '23 Fastcomposer (SD1.5) \cite{fastcomposer}              & 29.1     & 67.4& 42.3& 59.8  & '23 Perfusion (SD1.5)    \cite{tewel2023key}           & 23.2     & 41.6& 30.2 & 25.7         \\
  	   '23 PhotoMaker (SDXL) \cite{photomaker}   & 28.5       & 71.9& \underline{53.5}& 63.2   & '23 CustomDiffusion (SD1.5) \cite{custom-diffusion}   & 28.4       & 39.3& 31.4& 27.2     \\
            '24  InstantID  (SDXL)   \cite{wang2024instantid}              & 29.6      & 69.2& 51.8& 61.7   & '24 FreeCustom  (SD1.5)  \cite{freecustom}      & 31.4      & \underline{55.4}& \underline{42.1}& \underline{43.2}    \\

		'24 FlashFace (SD1.5) \cite{flashface}  & \underline{30.4}       & \underline{75.1}& 52.7& \underline{64.9}&  '24 ClassDiffuion (SD1.5) \cite{huang2024classdiffusion}  & \underline{32.8}       & 51.2& 38.5& 35.4    \\
  	
            \bottomrule
		\textbf{MagicFace} (SD1.5)              & \textbf{33.6} & \textbf{76.5}& \textbf{55.2}& \textbf{66.1}  & \textbf{MagicFace} (SD1.5)             & \textbf{34.2} & \textbf{59.4}& \textbf{45.3}& \textbf{51.8}  \\
		\bottomrule
	\end{tabular} \\

	\label{table:compare_methods}
\end{table*}

%% file: figs/multi-concept_compare1.tex
\begin{figure}[htb]
    \centering
    \includegraphics[width=1\linewidth]{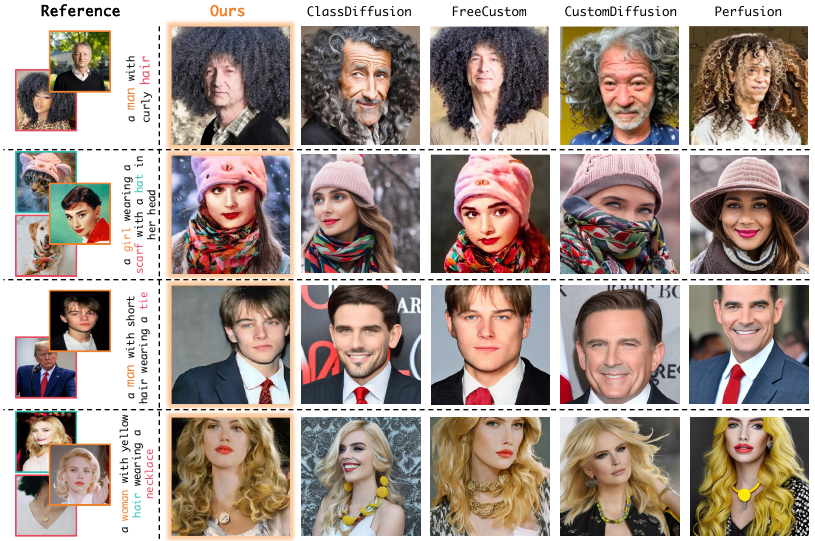}
    \caption{Qualitative comparison on multi-concept customization.}
    \label{fig:multi-concept_compare}
\end{figure}

%% file: figs/hyper_analysis1.tex
\begin{figure}[!htb]
    \centering
    \includegraphics[width=1\linewidth]{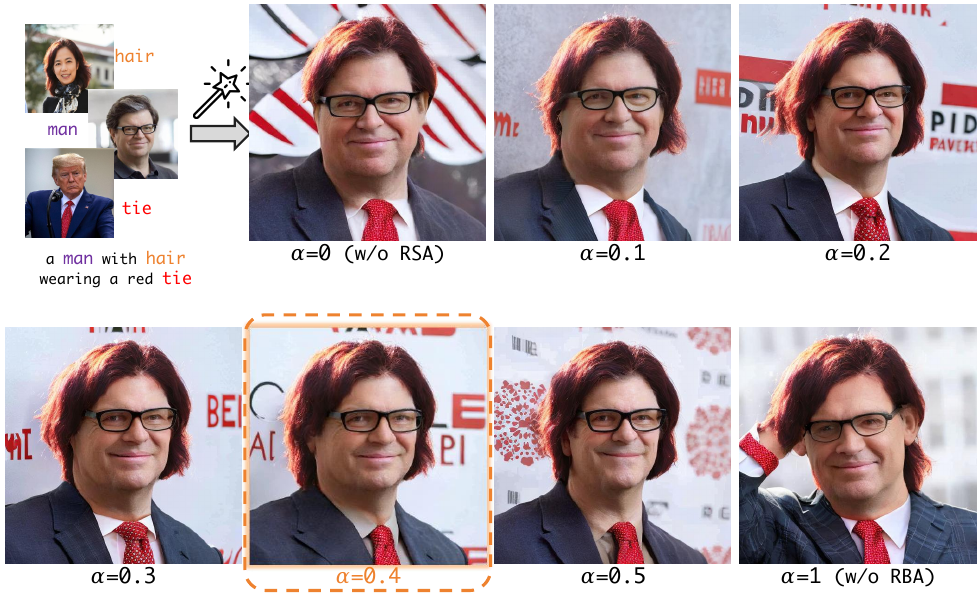}
    \caption{Hyperparameter analysis of $\alpha$.}
    \label{fig:hyper_analysis}
    \vspace{-0.5cm}
\end{figure}

%% file: tables/user_study_subject.tex
\begin{table}[t]
\setlength\tabcolsep{4pt}
\centering
\small
\caption{User study on subject-to-image generation: higher score, better ranking.}
\vspace{-0.2cm}
\begin{tabular}{ccccc}

		\bottomrule
		Methods
		& ID Fidelity       & Image Quality  &Prompt Consistency       \\
		\bottomrule

		PhotoMaker         &   2.28            & 2.06  &2.19        \\

		InstantID          &   2.07         & 2.23     & 2.21   \\

  		FlashFace       &    2.64      & 2.59  & 2.32   \\
        \bottomrule
            \textbf{Ours}          & \textbf{3.01}             & \textbf{3.12}    &\textbf{3.28}      \\
		\bottomrule
	\end{tabular} \\


\label{tab:user_study_subject}
\vspace{-0.3cm}
\end{table}

%% file: tables/user_study_multi.tex
\begin{table}[t]
\setlength\tabcolsep{2pt}
\centering

\small
\caption{User study on multi-concept customization: higher score, better ranking.}
\vspace{-0.2cm}
\begin{tabular}{ccccc}

		\bottomrule
		Methods
		& ID Fidelity       & Image Quality  &Prompt Consistency       \\
		\bottomrule

		CustomDiffusion           &   1.56         & 1.94     & 1.83   \\

  		FreeCustom      &    3.07      & 2.59  & 2.98   \\
      	ClassDiffusion      &    1.83      & 2.07  & 2.03   \\
        \bottomrule
            \textbf{Ours}          & \textbf{3.54}             & \textbf{3.40}    &\textbf{3.16}      \\
		\bottomrule
	\end{tabular} \\


\label{tab:user_study_multi}
\vspace{-0.3cm}
\end{table}

%% file: figs/appearance_transfer1.tex
\begin{figure}[hbt]
\vspace{-0.2cm}
    \centering
    \includegraphics[width=1.0\linewidth]{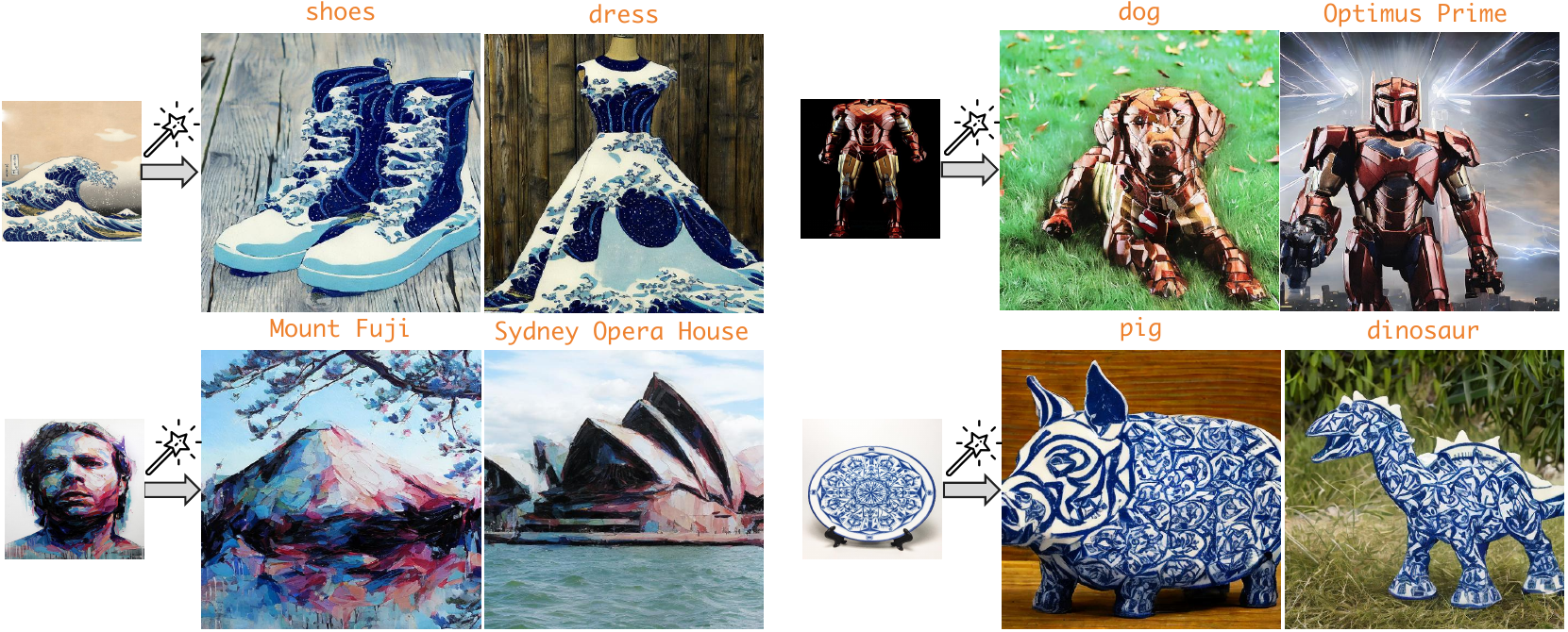}
    \caption{\textbf{Texture transfer}. Our method can inject the texture or materials of input images into generated objects.}
    \label{fig:appearance_transfer}
    \vspace{-0.4cm}
\end{figure}

%% file: 10_conclusion.tex
\section{Conclusion}
\label{sec:conclusion}

In this paper, we propose MagicFace, the first universal-style human image personalization method that enables multi-concept customization. Inspired by the human drawing process, our approach first outlines a semantic layout by considering coarse elements such as concept shape and pose, and then refines the image by incorporating fine-grained features of each concept. To achieve this, we propose a coarse-to-fine generation pipeline with two stages: RSA (Reference-aware Self-Attention) for establishing an initial semantic layout by extracting semantic information from all reference images, and RBA (Reference-grouped Blend Attention) for refining this layout by accurately injecting fine-grained features from each reference concept into its corresponding latent regions.
Extensive experiments highlight its superiority both qualitatively and quantitatively. 

%% file: 12_appendix.tex
\appendix
\label{sec:appendix}

\input{figs/more_compare1}

\section{Superiority compared with baselines}
\textbf{Training-Free Approach.} Existing human-centric subject-to-image generation methods typically rely on extensive retraining on large-scale datasets or fine-tuning with dozens of images. These approaches involve time-consuming processes, making rapid deployment challenging. In contrast, MagicFace is entirely training-free, eliminating the need for large-scale pre-training and the associated computational overhead. By requiring only a single image per concept, our approach is significantly more efficient and practical, reducing both time and computational resource demands.
\noindent\textbf{High-fidelity Results.} Despite the simplicity, our method consistently delivers more natural and realistic human personalization results. Extensive quantitative and qualitative evaluations demonstrate that our approach matches or even surpasses the performance of more complex, training-based methods, highlighting its effectiveness in producing high-fidelity human images.

\noindent\textbf{Versatility in Applications.} 
(1) \textit{Universal-style human customization.} 
Unlike existing methods constrained by their training datasets to only photorealistic styles, our method excels at customizing a wide range of styles. 
By accurately embedding reference concept features into the generated image in an evolving scheme, our method is adept at customized synthesizing images of humans across diverse styles.
(2) \textit{Texture transfer.} Our approach is not only superior in human image synthesis but also highly effective for texture transfer. By precisely extracting appearance features from input images and seamlessly integrating these features into generated objects, our method shows its robustness across different applications.

\noindent\textbf{Multi-concept Human Customization.} 
Current subject-to-image generation methods centered on humans struggle with multi-concept customization, falling short in accurately personalizing individuals with multiple given attributes. In contrast, our approach achieves high-quality, multi-concept human customization, providing an advanced level of flexibility and precision.
To ensure a comprehensive evaluation, we also compared our method against specialized baselines in the multi-concept customization field. While these existing methods are fairly effective for general objects with coarse-grained textures, they consistently fall short in human-centered customization tasks. Our experimental results demonstrate that MagicFace excels in preserving human identity, establishing a new standard in multi-concept human image synthesis.

\section{More compared baselines}
We also compare our method with tuning-based baselines, i.e., Dreambooth \cite{dreambooth} and Text Inversion \cite{textual_inversion}, and a zero-shot baseline, IP-adapter \cite{ye2023ip}. The quantitative and qualitative results are provided in \cref{table:compare_supp} and \cref{fig:more_compare}, respectively. 

\input{tables/compare_supp}

\section{Inference time comparison}
Our method is entirely training-free and personalizes a human image with just a single forward pass. We also compare inference times with selected efficient baselines using 50 steps of the DDIM sampler, as shown in \cref{tab:time_subject}.

\input{tables/compare_time_subject}

\section{More visual results}
More visual results generated by our method are shown in \cref{fig:more_results1} and \cref{fig:more_results2}.

\section{Choice of self-attention layer replacement}
We explore the optimal choice of replacing the original self-attention in the basic block with our RSA/RBA, as shown in \cref{fig:choice_replacement1} and \cref{fig:choice_replacement2}.  The results indicate that replacing the self-attention layers in blocks 5 and 6 produces the highest fidelity images.

\section{Choice of weight $w$}
We provide more cases for exploring the impacts of different weight settings in \cref{fig:weighted_mask_supp}.

\section{Choice of hyperparameter $\alpha$}
We provide an additional case for exploring the optimal value of $\alpha$ as shown in \cref{fig:hyper_analysis_supp}.

\input{figs/hyper_analysis2}

\section{Visualizatoin of RSA and RBA}
More visualization results of RSA and RBA are shown in \cref{fig:visualized_attention_supp}.


\input{figs/more_results1}
\input{figs/more_results22}
\input{figs/visualized_attention_supp1}
\input{figs/weight_choice_supp1}
\input{figs/choice_replacement11}
\input{figs/choice_replacement22}

%% file: figs/more_compare1.tex
\begin{figure*}[hbt]
    \centering
    \includegraphics[width=1\linewidth]{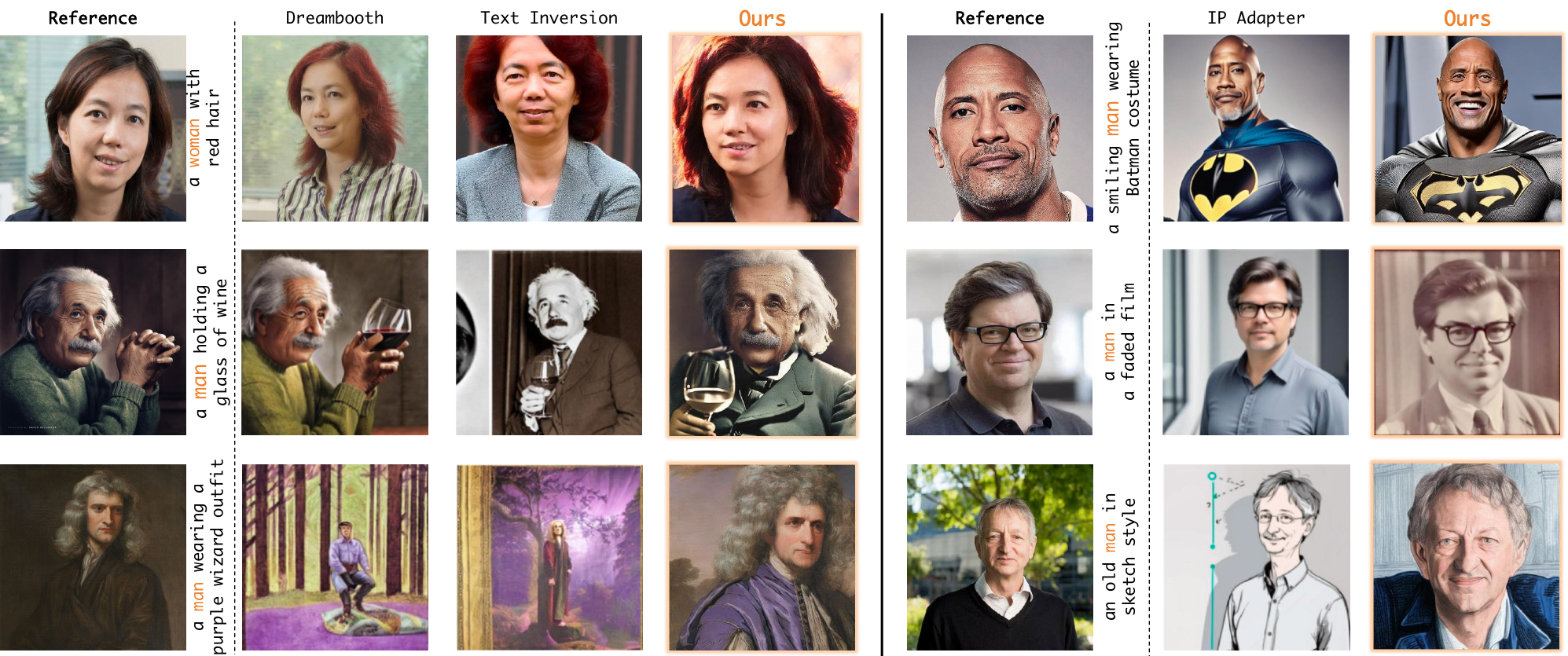}
    \caption{Qualitative comparative results on subject-to-image generation.}
    \label{fig:more_compare}
    \vspace{-0.4cm}
\end{figure*}

%% file: tables/compare_supp.tex
\begin{table}[ht]  \centering
	\small
        \setlength{\tabcolsep}{3pt}
	\caption{Quantitative comparison against baselines.}
	\label{table1}
	\begin{tabular}{c|cccc}
		\bottomrule
		\multirow{2}{*}{\textbf{Methods}} & \multicolumn{4}{c}{\textit{\textbf{Subject-to-Image Generation}}} \\
		  
		& CLIP-T      & CLIP-I     &DINO & Face-sim.        \\
		\bottomrule

            '23 Dreambooth (SDXL)               & 31.6     & 58.4 & 37.4& 43.2        \\
  	   '23 Text Inversion (SD1.5)    & 26.7       & 63.1& 36.9& 48.6    \\
      
		'23 IP-adapter (SDXL)  & 27.3       & 64.9& 47.5& 62.8  \\
  	
            \bottomrule
		\textbf{MagicFace} (SD1.5)              & \textbf{33.6} & \textbf{76.5}& \textbf{55.2}& \textbf{66.1}  \\
		\bottomrule
	\end{tabular} \\

	\label{table:compare_supp}
\end{table}

%% file: tables/compare_time_subject.tex
\begin{table}[t]
\setlength\tabcolsep{3pt}
\centering
\small
\caption{Inference time comparison. '-' indicates that the information is not available.}

\begin{tabular}{ccccc}

		\bottomrule
		Methods & Fastcomposer    & Photomaker  & FreeCustom  & Ours   \\
		\bottomrule
            single-concept     & \textbf{7s}  & 12s   & 20s  &  10s \\
            2-concept     & -  & -   & 36s   & \textbf{18s} \\
            3-concept     & -  & -  & 58s   & \textbf{25s} \\
		\bottomrule
	\end{tabular} \\

\label{tab:time_subject}
\vspace{-0.3cm}
\end{table}

%% file: figs/hyper_analysis2.tex
\begin{figure}[htb]
    \centering
    \includegraphics[width=1\linewidth]{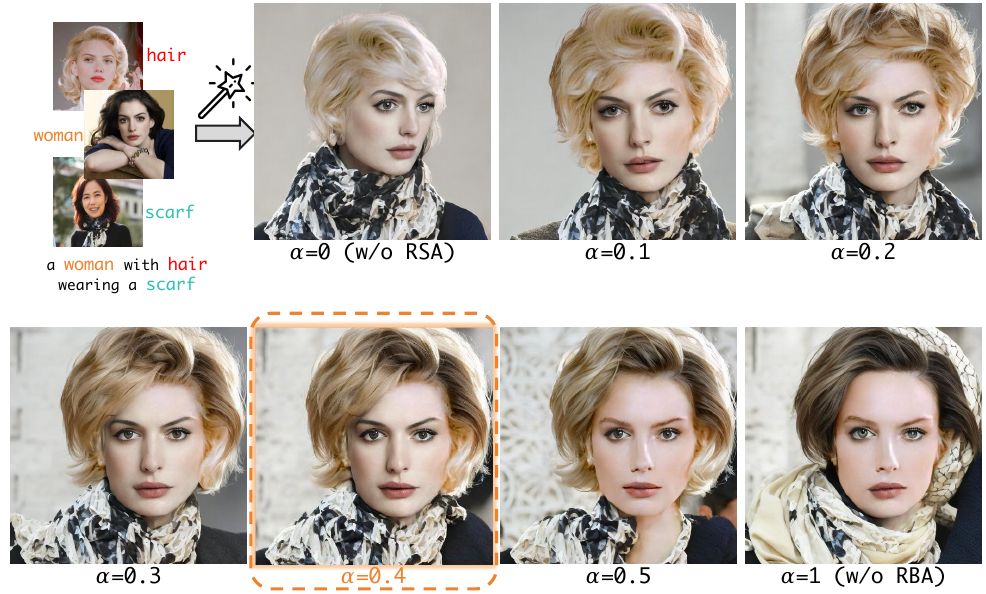}
    \caption{Hyperparameter analysis of $\alpha$.}
    \label{fig:hyper_analysis_supp}
\end{figure}

%% file: figs/more_results1.tex
\begin{figure*}[hbt]
    \centering
    \includegraphics[width=1\linewidth]{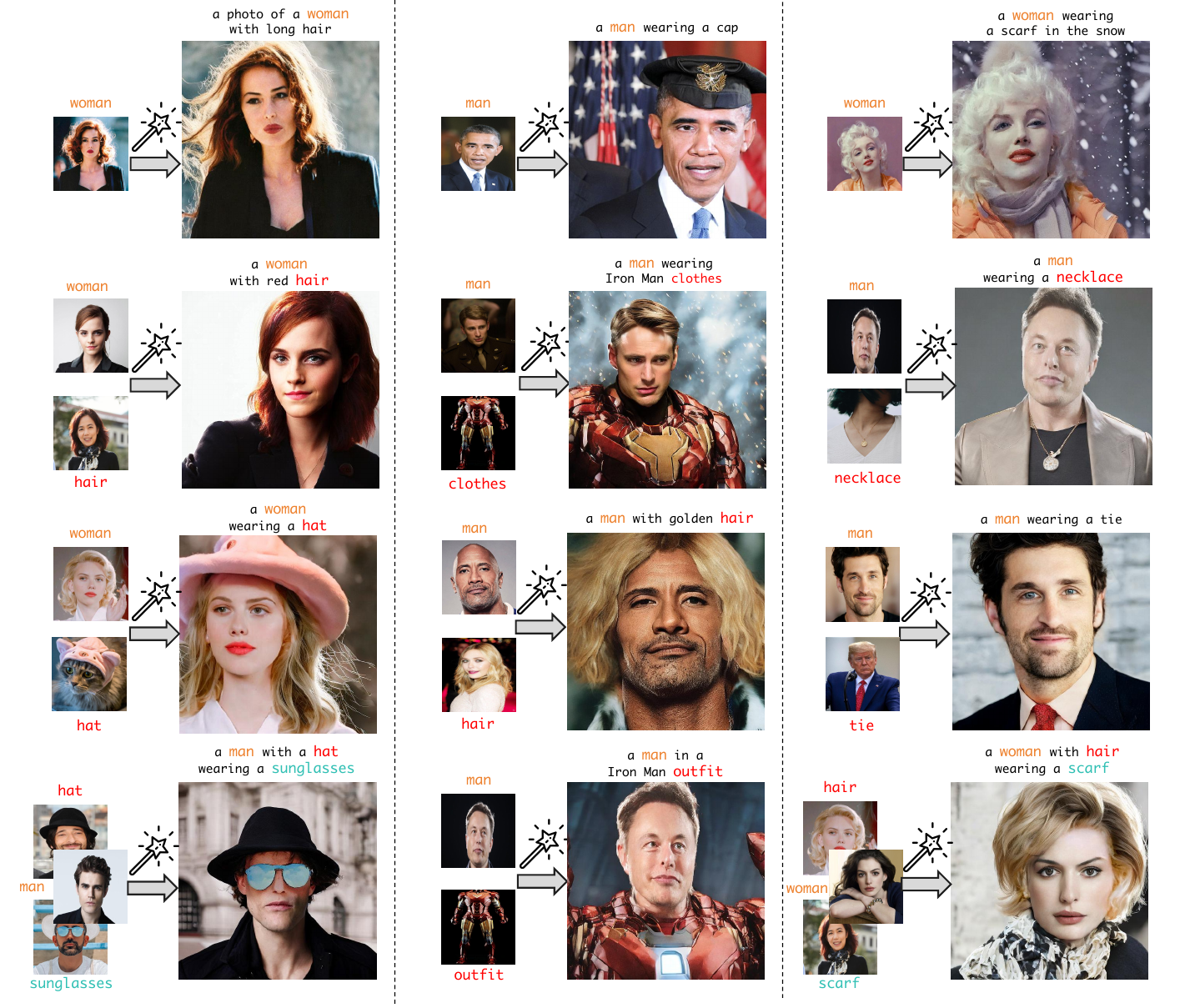}
    \caption{More visual results of single/multi-concept customization for humans of photorealism style.}
    \label{fig:more_results1}
\end{figure*}

%% file: figs/more_results22.tex
\begin{figure*}[hbt]
    \centering
    \includegraphics[width=1\linewidth]{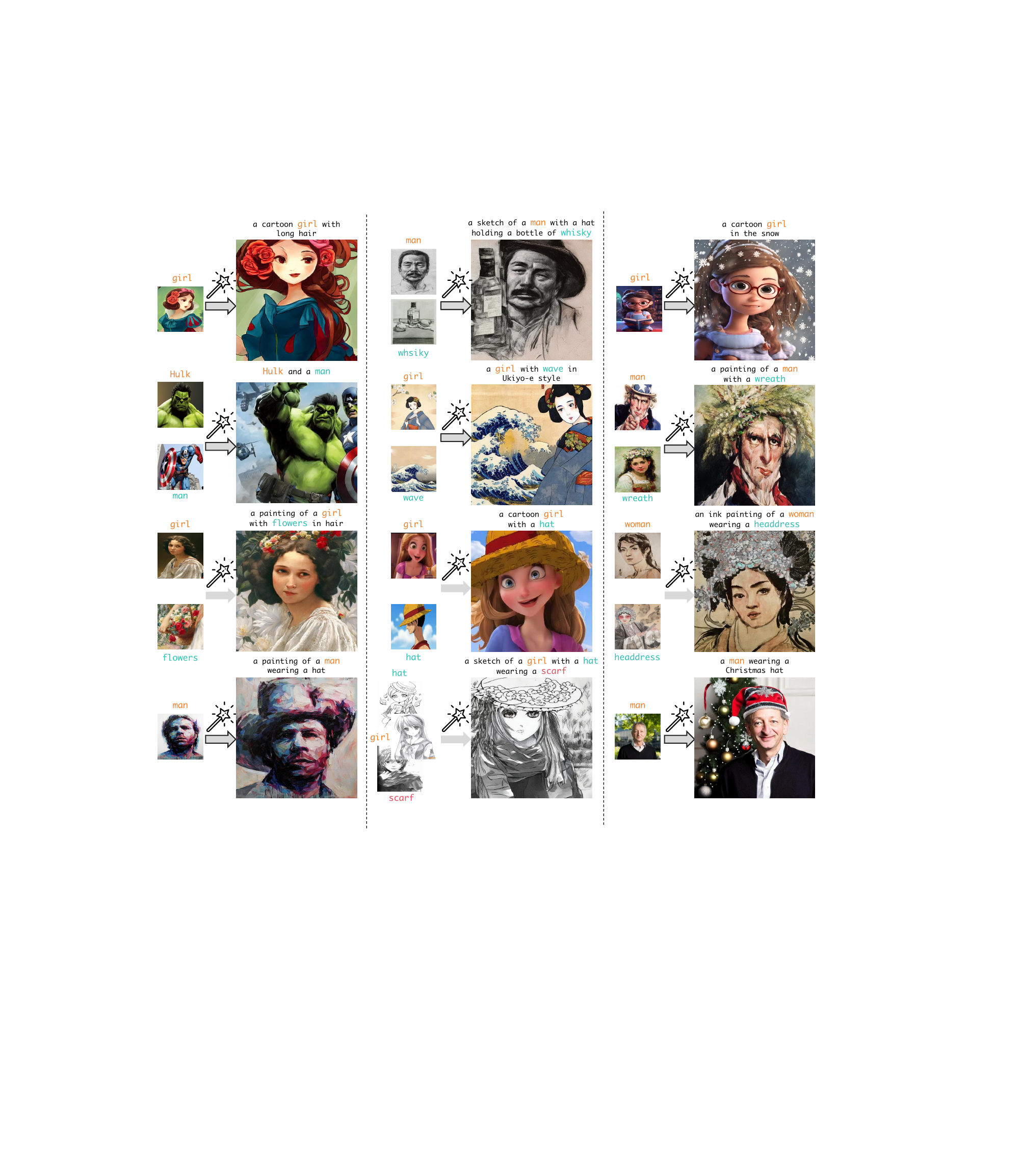}
    \caption{More visual results of single/multi-concept customization for humans of various styles.}
    \label{fig:more_results2}

\end{figure*}

%% file: figs/visualized_attention_supp1.tex
\begin{figure*}[htb]
    \centering
    \includegraphics[width=0.7\linewidth]{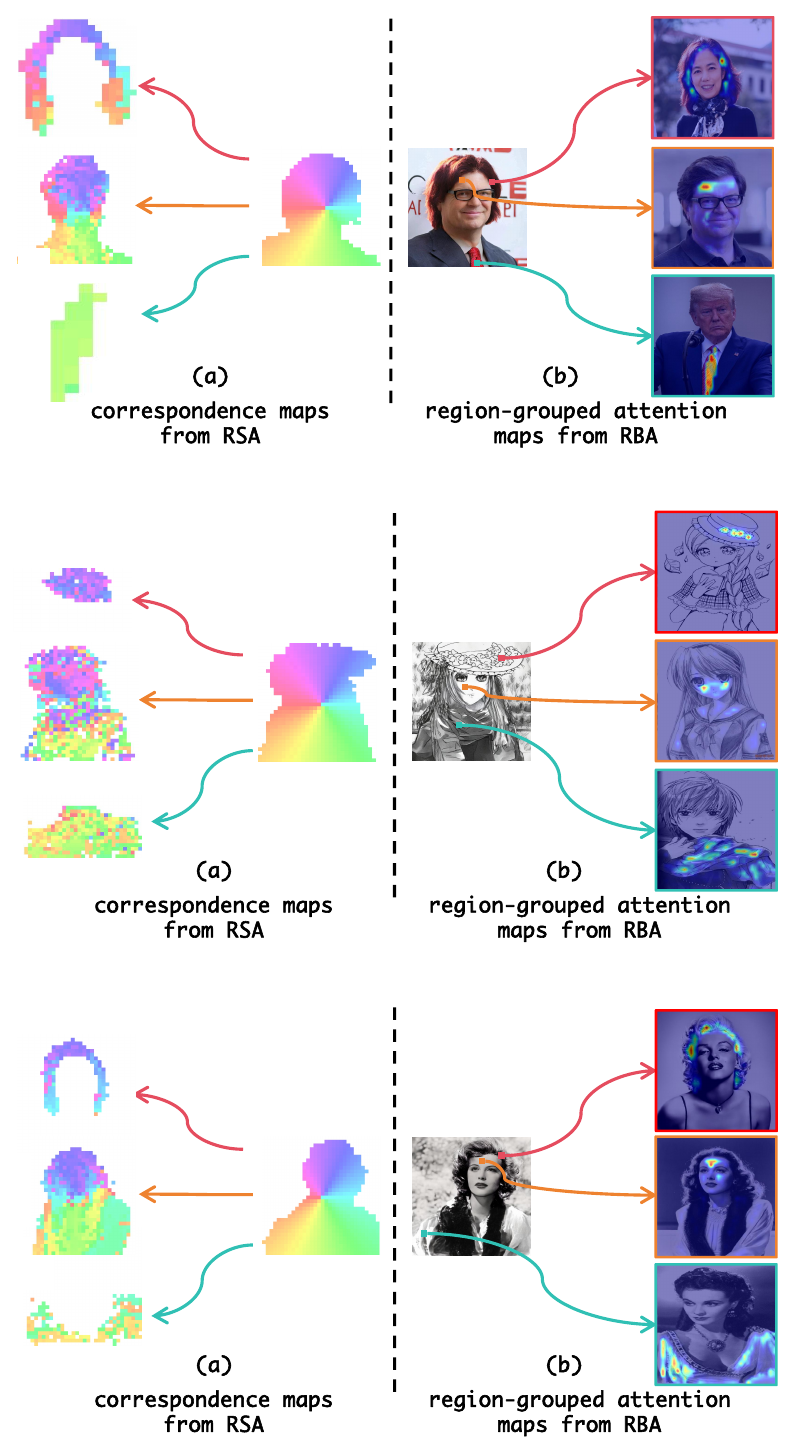}
    \caption{\textbf{Correspondence maps and region-grouped attention maps visualization.} In (a), features with the highest similarity between the generated subject and the reference concepts are marked with the same color. (b) the results of features in colored boxes querying their reference concept keys. }
    \label{fig:visualized_attention_supp}
\end{figure*}

%% file: figs/weight_choice_supp1.tex
\begin{figure*}[t]
    \centering
    \includegraphics[width=0.9\linewidth]{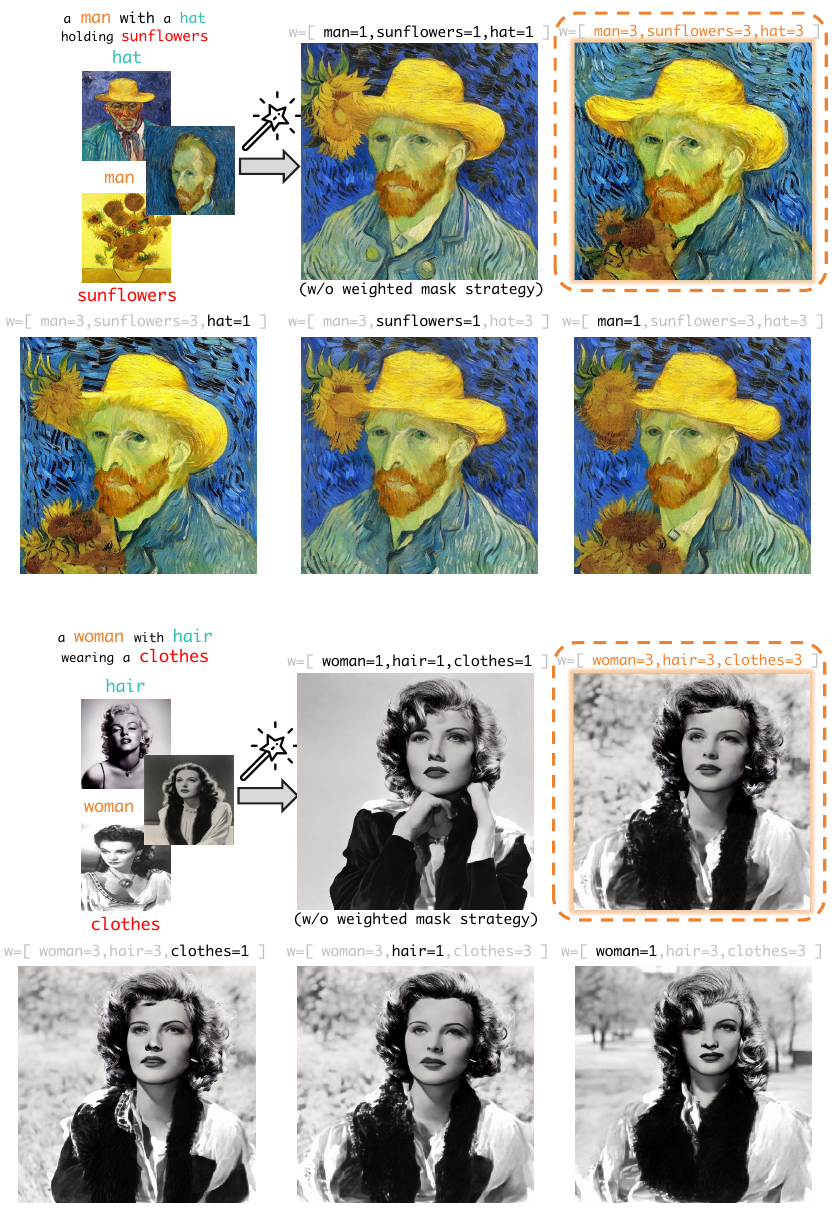}
    \caption{Visualized results under different weight settings $w$.}
    \label{fig:weighted_mask_supp}
\end{figure*}

%% file: figs/choice_replacement11.tex
\begin{figure*}[t]
    \centering
    \includegraphics[width=1\linewidth]{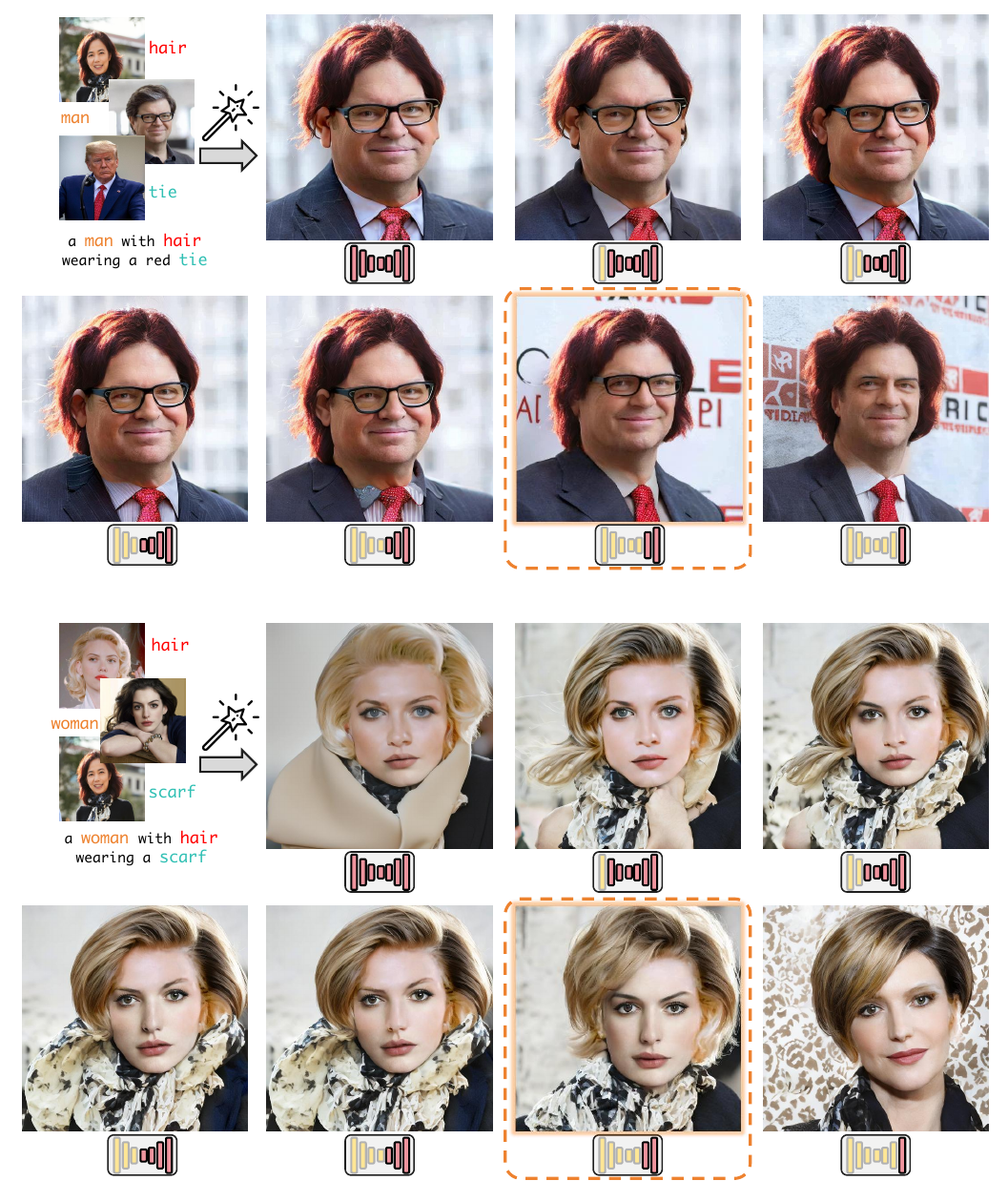}
    \caption{\textbf{Choice of self-attention layer replacement}. The yellow color represents the original basic block, while the red color indicates the basic block where the self-attention modules have been replaced by RSA/RBA.}
    \label{fig:choice_replacement1}
\end{figure*}

%% file: figs/choice_replacement22.tex
\begin{figure*}[t]
    \centering
    \includegraphics[width=1\linewidth]{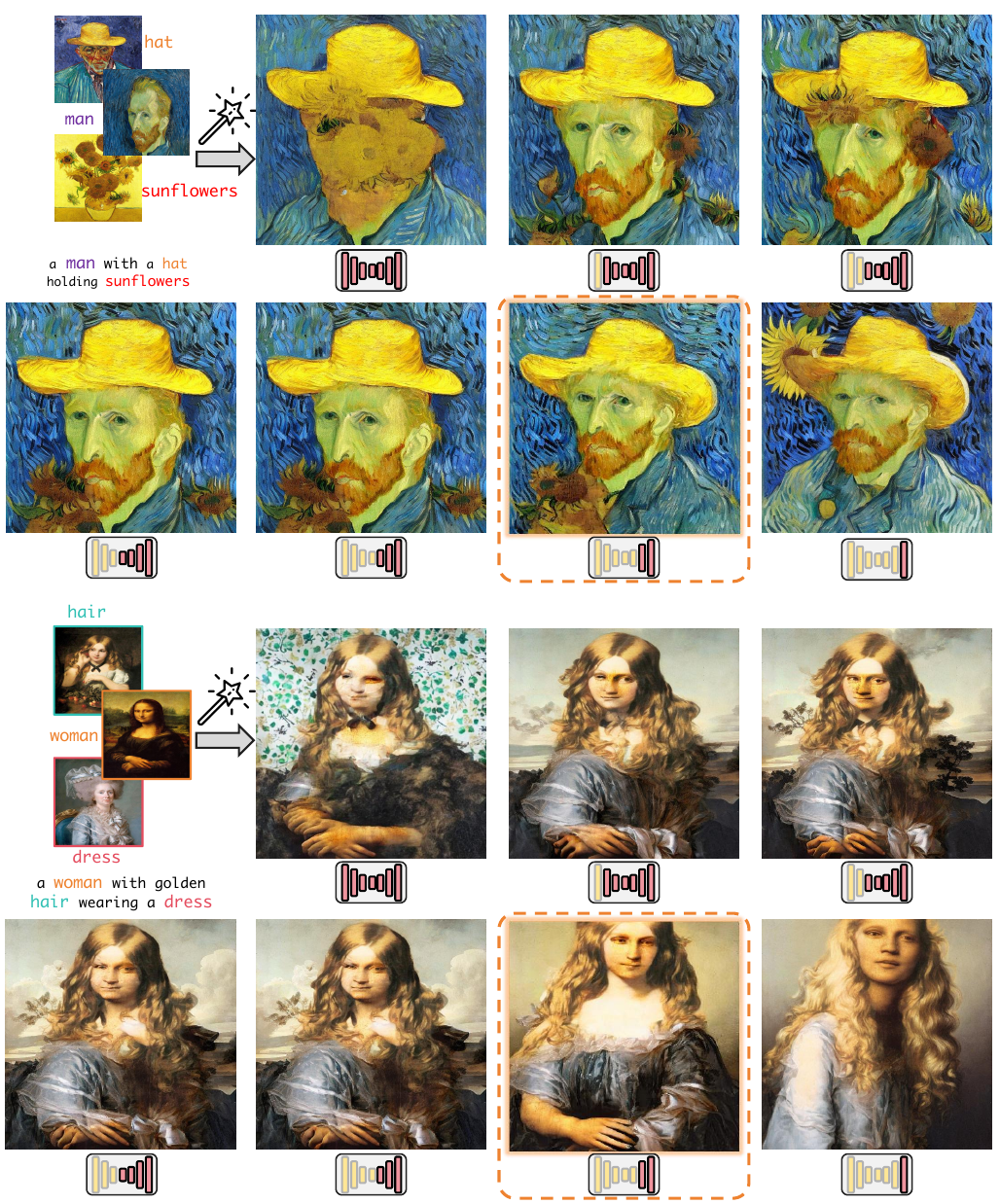}
    \caption{\textbf{Choice of self-attention layer replacement}. The yellow color represents the original basic block, while the red color indicates the basic block where the self-attention modules have been replaced by RSA/RBA.}
    \label{fig:choice_replacement2}
\end{figure*}